# Health improvement framework for planning actionable treatment process using surrogate Bayesian model

Kazuki Nakamura[1,2], Ryosuke Kojima[2], Eiichiro Uchino[2], Koichi Murashita[3], Ken Itoh[4], Shigeyuki Nakaji[5] and Yasushi Okuno[2*]

[1] Research & Business Development Department, Kyowa Hakko Bio Co., Ltd., Tokyo, Japan.

[2] Department of Biomedical Data Intelligence, Graduate School of Medicine, Kyoto University, Kyoto, Japan.

[3] Center of Innovation Research Initiatives Organization, Hirosaki University, Zaifu-cho, Hirosaki, Aomori, Japan.

[4] Department of Stress Response Science, Hirosaki University Graduate School of Medicine, Zaifu-cho, Hirosaki, Aomori, Japan.

[5] Department of Social Health, Hirosaki University Graduate School of Medicine, Zaifu-cho, Hirosaki, Aomori, Japan.

## Abstract

Clinical decision making regarding treatments based on personal characteristics leads to effective health improvements. Machine learning (ML) has been the primary concern of diagnosis support according to comprehensive patient information. However, the remaining prominent issue is the development of objective treatment processes in clinical situations. This study proposes a novel framework to plan treatment processes in a data-driven manner. A key point of the framework is the evaluation of the "actionability" for personal health improvements by using a surrogate Bayesian model in addition to a high-performance nonlinear ML model. We first evaluated the framework from the viewpoint of its methodology using a synthetic dataset. Subsequently, the framework was applied to an actual health checkup dataset comprising data from 3,132 participants, to improve systolic blood pressure values at the individual level. We confirmed that the computed treatment processes are actionable and consistent with clinical knowledge for lowering blood pressure. These results demonstrate that our framework could contribute toward decision making in the medical field, providing clinicians with deeper insights.

* Address correspondence to: Kazuki Nakamura (nakamura.kazuki.88m@st.kyoto-u.ac.jp) or Yasushi Okuno (okuno.yasushi.4c@kyoto-u.ac.jp).

# Introduction

Medically appropriate and patient-acceptable decision making is beneficial in enhancing the quality of care[1–3]. Considerable evidence has been accumulated from many studies on the assessment of health statuses and risk profiles. Accordingly, standardized care has been provided in the form of guidelines. Conversely, the uniform application of standardized care is undesirable in real clinical situations whereby diversity exists in relation to individual preferences, feasibility, and acceptability[4]. Personalized medicine, also known as precision medicine, has become popular and presents new opportunities in clinical situations in recent years[5–7]. In addition to the understanding of individual, unique health-related factors, their consideration has been essential for shared decision making between patients and clinicians. Realistic and appropriate health improvement plans for the patient's health conditions have been regarded as a crucial component of shared decision making[8,9]. However, feasible planning largely depends on the empirical judgment of the clinicians in clinical situations, and decision making based on objective health improvement plans for both patients and clinicians is difficult. Major challenges remain in providing clinicians with tools to support clinical decision making in a data-driven manner[10,11].

Machine learning (ML) technology has been extensively used in the medical field, especially for diagnosis support and disease prediction based on comprehensive patient information[12–14]. Unlike interpretable techniques, such as classical statistical analysis and linear models, the black-box nature of highly predictive ML models, including ensemble learning and deep learning, is often a barrier to clinical decision making applications[15]. Explainable artificial intelligence (XAI) has been receiving increasing attention recently in the field of ML[16]. XAI is a research field on techniques that explain black-box ML predictions. A successful application of XAI in the medical field is the identification of individual, health-related factors that contribute to disease prediction using Local Interpretable Model-agnostic Explanations (LIME) and SHapley Additive exPlanations (SHAP)[17–22]. These methods achieve both predictive performance and individual interpretation by using an additional individual model referred to as the surrogate model. The remaining important issue in clinical decision making is the development of personalized treatment plans for rational treatment[23]. However, these conventional methods merely explain the prediction reasons but cannot provide effective treatment processes. For example, for the prediction of hypertension, it is unclear what type of actions will improve effectively the individual blood pressure in a set of candidate actions related to blood test data and body composition despite the fact that we can understand important features in the prediction.

This study proposes a novel framework for planning an actionable path for personalized treatment based on the predictions of a ML model. A key idea of our framework is to use a hierarchical Bayesian model as a surrogate model of a specified ML model. We refer to this surrogate model as the *stochastic surrogate model*. Our framework can evaluate the "actionability" of the treatment processes that was not considered by conventional methods[17,18], by computing a path probability with the use of the stochastic surrogate model. The combined use of the ML model and the stochastic surrogate model achieves both a high-prediction performance and actionability evaluation. The simultaneous computation of ML model prediction and actionability evaluation yields an actionable treatment process that leads to clinical applications to improve personal health.

This study also presents experiments conducted to address two different aspects. First, we evaluated our proposed framework from the viewpoint of the proposed methodology using a synthetic dataset related to a regression task. Subsequently, we assessed our framework on a clinical application using an actual health checkup dataset. In this experiment, we calculated personal health improvement paths and confirmed the consistency with clinical knowledge based on the assumption of a scenario wherein we aimed to lower blood pressure in individuals with high-blood pressure. To the best of our knowledge, this is the first study that made it possible to present effective improvement paths based on the ML model using an actual health checkup dataset.

# Results

## Path planning framework using surrogate Bayesian model

This section describes the proposed framework (Fig. 1, Supplementary Fig. 1). Our framework consisted of three steps. Step 1: We built a regression model using ML methods. Step 2: The surrogate model of the regression model was constructed using hierarchical Bayesian modeling. Step 3: Path planning was conducted to identify an actionable path for the treatment performed using the surrogate model. Thus, the combined framework of the regression model and the stochastic surrogate model achieves both high prediction performance and actionability evaluation, which leads to an actionable treatment process for personal health improvement.

In Step 1, a regression model was built from a dataset expressed in the form of a table format, wherein columns consist of multiple explanatory variables and a response variable, and rows represent instances (Fig. 1a). The output of the constructed model would correspond to a clinician's assessment of individual health status or future predictions. For example, a regression model for blood pressure could be used to estimate the value of the response variable, i.e., blood pressure, from the value of explanatory variables, such as the body composition and blood test data. In our framework, we can use arbitrary algorithms, such as high-performance nonlinear algorithms, to construct the regression model.

Based on the original explanatory variable values and the prediction values of the regression model in Step 1, a stochastic surrogate model was constructed in Step 2 using hierarchical Bayesian modeling (Figs. 1b and 2). Items that cannot be easily measured in clinical situations, or future values, are available as the response variables given that the prediction values of the ML model were applied in our framework. Clinically, this stochastic surrogate model represents a set of realistic health conditions for patients. The model was used to compute the probability of given values of explanatory and response variables to evaluate actionability in the next step. Note that this stochastic surrogate model represents a probability density on not only a given dataset but also virtually changed values, i.e., counterfactual values.

In Step 3, an optimal health-improvement treatment path was calculated for each instance using the surrogate model (Figs. 1c, 1d, and 3, as detailed in the methods section), and would be equivalent to the consideration of an appropriate health improvement plan for individual patients in clinical situations. The explanatory variables of an instance, such as body composition and blood test data, were hypothetically changed to improve the response variable predicted using the regression model. By using the surrogate model constructed in Step 2, we could calculate the probability of counterfactual values. In our framework, the optimal path was defined as a sequence of the counterfactual values with high probability in the surrogate model. Therefore, this model can avoid the nonoptimal health-improving paths (shown by the red line in Fig. 1c) where intermediate situations may be unrealistic.

Considering real applications, explanatory variables contain variables that cannot be changed by interventions such as age. The appropriate subset of variables for the application should be determined. The explanatory variables for setting the counterfactual values were called intervention variables in this study. In our path planning setting, the intervention variables were regarded as a grid graph, and a path was defined by connecting the grid points (nodes). We defined a probability of the node as the probability of taking the node calculated using the surrogate model. Furthermore, the actionability was defined as the product of node probabilities on a specified path. Based on a breadth-first search, we obtained the most actionable path to the destination node that achieves the most improved predictive value within the search iteration count $L$ (detailed in the methods section).

**Validation of framework on synthetic dataset**

We evaluated our framework with a synthetic dataset to confirm that improvement paths with high actionability can be planned. This synthetic dataset was generated from three, three-dimensional (3D) normal distributions (Supplementary Fig. 2). We built a regression model and subsequently constructed a surrogate model using hierarchical Bayesian modeling (Supplementary Figs. 3a and 3b). The lowest widely applicable Bayesian information criterion (WBIC) value[24] was obtained when the number of mixture components in the hierarchical Bayesian model was equal to two. Details of the setting related to this model are given in the methods section. Subsequently, we planned paths to decrease the value of the response variable using this surrogate model. All explanatory variables were selected for intervention variables, and planned paths were more actionable than the baseline path (Supplementary Fig. 3c). This baseline method for the baseline path is defined in the methods section. The planned paths of two randomly selected instances are shown in Fig. 4. We successfully demonstrated that our framework could plan paths to improve response variable values with high probabilities in our framework. These results showed that the planned paths with high probabilities were discovered rather than the naïve straight path that connected the initial node to the destination node.

**Application on actual health checkup dataset**

We used the Iwaki Health Promotion Project (IHPP) dataset, an actual health checkup dataset, to demonstrate that the proposed framework can plan actionable paths for treatment. The IHPP has acquired (on an annual basis) a wide range of health checkup data that describe the molecular biology, physiology, biochemistry, lifestyle, and socio-environment of participants. We considered a scenario to improve systolic blood pressure (SBP) whereby planned paths could be interpreted from a clinical perspective. Table 1 shows an overview of the IHPP dataset. Because the dataset comprised more than 2,000 measurement items, we reduced the explanatory variables before we built a regression model in our framework. We excluded measurement items related to blood pressure from the explanatory variables. Furthermore, ambiguous items, such as answers to questionnaires and items with ≥25% missing values, were excluded. Subsequently, XGBoost-based[25] recursive feature elimination (RFE)[26] was performed to reduce the explanatory variables with the training data. We applied one-hot encoding for categorical variables and replaced the missing values with the median for simplicity. RFE was performed with five-fold cross-validation split by participants, and the explanatory variables were reduced to 25 (Fig. 5a, the details of the variables are described in Supplementary Table 1). Important features comprised items related to hypertension, such as age, body composition (leg score, body mass index (BMI), and waist), blood glucose, and gamma glutamyl transferase (γ-GTP)[27–35]. Therefore, the selected explanatory variables were considered to be reasonable for SBP predictive models from the clinical perspective.

Following our framework, a regression model was built after the replacement of missing values with the use of multiple imputations. This is a more precise imputation method for missing values. To estimate multiple imputations, Bayesian ridge and random forest[36] were used for continuous and discrete variables, respectively. Consequently, the regression model yielded a root-mean-squared error (RMSE) equal to 15.42 and an R-squared value equal to 0.330 (Fig. 5b). Subsequently, hierarchical Bayesian modeling was performed to construct the surrogate model. The lowest WBIC value was obtained when the number of mixture components was five (Fig. 5c).

Path planning was performed with the surrogate model. Regarding the intervention variables, the top-five variables that could be intervened were selected based on feature importance: leg score, blood glucose, BMI, waist, and γ-GTP. The unit cell size of the grid was set to $0.2\ \sigma$ for each explanatory variable, where $\sigma$ represents the standard deviation of the training data. Assuming a scenario wherein the task is to lower the SBP in participants with higher values, relevant instances were selected according to the following criteria: SBP above the mean + $(1-\sigma)$, and no missing values in the intervention variables. The number of applicable instances was 391. We executed the path-search algorithm with $L = 20{,}000$ for each instance and acquired a path with the lowest predictive SBP value.

For quantitative evaluations, we introduce the actionability score for each instance that indicates how actionable the planned path was compared with the baseline path (detailed in the methods section). The histogram of the actionability score is shown in Fig. 5d. The actionability scores were greater than zero, i.e., planned paths were more actionable in 341/391 instances, and the median was 0.78. This result suggested that even if the response variable value after improvement was the same, the path planned by our framework had a higher actionability than the baseline path in most cases.

In the subsequent part, we show the paths of the three randomly selected instances (Fig. 6). The actual data scatters also support the fact that the path was planned to pass through areas with high-nodal probabilities. In instance 1, it was shown that the path that improved the values of the variables in the following order was more effective: blood glucose, leg score, and γ-GTP (Figs. 6a and 6b). These variables are related to each other, and the path in which multiple variables fluctuated to improve the blood pressure was reasonable[37–39]. In instance 2, the path that improved the values in the following order was more valid: γ-GTP, leg score, and again γ-GTP (Figs. 6c and 6d). The SBP values predicted by the regression model increased temporarily compared with the original SBP value. In instance 3, the optimal path was planned only based on the improvement of one explanatory variable, namely, the leg score (Figs. 6e and 6f). The actionability score yielded a value of zero because the optimal path was identical to the baseline path in such cases.

## Discussion

In this study, we proposed a framework for planning paths to improve the prediction values of ML models. We demonstrated that the proposed framework could plan paths through nodes with high probabilities using the synthetic dataset. Furthermore, our proposed framework was capable of planning actionable paths to improve the predicted SBP values in the actual health dataset. Our framework suggests realistic concrete treatment processes that are actionable on humans for personal health improvement. Conventional XAI methods, such as LIME and SHAP, cannot provide concrete improvement paths.

As shown in Fig. 6, our framework could visually present concrete improvement paths at the individual level. The paths mainly consisted of a sequence of changes in leg score, blood glucose, and γ-GTP among the intervention variables. The direction of change in these variables was consistent with conventional clinical knowledge for improving blood pressure[29–35]. For example, high-blood glucose levels have been reported to be a risk factor of hypertension. Tool-assisted goal-settings are expected to help time-constrained clinicians and contribute to better health improvement in patients[40]. Our framework can provide clinicians with understandable and acceptable health improvement plans based on patient health data and given intervention variables. Accordingly, this is suitable for patient–clinician collaborative decision making on health interventions.

In the remaining part of this section, we provide methodological considerations of the proposed framework. Although the experiments were performed under specific settings, our framework has generality in terms of its methodology owing to three aspects. First, we used XGBoost[25] to build the regression model to be explained. Because our framework operates in a model-agnostic manner, other high-performance regression models, such as deep learning, can be used. Second, although we assumed normal or categorical distributions for the explanatory variables in the hierarchical Bayesian modeling (Fig. 2), distributions can be selected according to the data. This is expected to be applied to some extent to medical data, which is often accompanied by a significant amount of noise and missing values. Finally, we selected intervention variables based on the feature importance in Step 1 of the experiment on the health checkup data. Also, in Step 3, the unit cell size in planning was set to 0.2 σ of the data distribution. These selection methods and values can be flexibly changed according to the application or the patient's request and environment.

Our objective in path planning was to obtain a path to the best-predicted value with the set conditions of the number of iterations $L$. In real situations, application-dependent or clinical constraints may exist. A typical case is that the values of explanatory and/or response variables should be less/more than the reference values in all nodes along the path. For example, there were cases where the predicted SBP values were temporarily increased from the original value in our experiment (Fig. 6c). This might be better avoided in clinical situations. By slightly changing the path search condition to exclude undesirable nodes, our framework can be applied to these cases. Additionally, when the target value of the response variable is determined based on guidelines or clinical knowledge, our framework can be applied by modifying the termination conditions of the search to reach the target value.

From the perspective of expanding the proposed framework, the high-computational cost when many intervention variables exist or when calculating a long-term path for treatment needs to be considered. Subject to our experimental conditions, approximately 10 min were required for path planning per instance. A more practical path planning can be expected by combining our framework with techniques for finding intervention points, such as counterfactual explanations[41–45]. Counterfactual explanations usually present intervention goal values of the explanatory variables for changing the response variable without considering the intervention process. Our framework can plan actionable paths to the intervention goal values decided by counterfactual explanations. For this case, a more efficient path planning algorithm, such as the A* search algorithm, can be applied in our framework. Similar to our actionable path planning approach, some studies have been conducted in recent years to obtain intervention points with minimum costs[45]. These approaches enable fast searches by assuming linearity. Our framework is more suitable for use with sophisticated nonlinear ML models, i.e., in cases where linearity is difficult to assume, such as those pertaining to medical checkup data.

Our study has some limitations. First, the health checkup dataset used in this study was obtained from a single area and had a small sample size. This may have contributed to the low-prediction score of the SBP regression model (Fig. 5b). Although the dataset problem does not impair the validity of the proposed framework as a methodology, we performed supplementary experiments that apply the framework on the public datasets to support the framework's validity (Supplementary Information). Furthermore, we encountered a problem when we verified the clinical effectiveness of the paths planned by using our framework. Planned paths in the synthetic dataset indicated that our framework would perform correctly. Although we evaluated that the change directions of intervention variables in some paths were consistent with clinical knowledge in the clinical application, it is necessary to verify the effectiveness of the paths through a prospective cohort study to suit the real-world applications of our framework.

In conclusion, we proposed a novel framework to plan actionable health improvement processes at the individual level. Using the synthetic dataset, we proved that our framework could plan actionable paths through the nodes with high probabilities. Furthermore, we successfully demonstrated that health-improving paths planned for lowering blood pressure based on the application of our framework to the actual health checkup dataset were actionable and consistent with clinical knowledge. Our framework can present reasonable and personalized health improvement plans based on ML model predictions in a wide range of situations that is expected to contribute to decision making in the medical field. Further studies should focus on the prospective clinical validation of actionable paths planned by using the framework proposed herein. Our framework may provide clinicians with deeper insights by proposing definite and actionable treatment paths through the use of the ML model.

## Materials and methods

### Synthetic 3D dataset

We generated a simple 3D dataset to verify whether our framework can plan paths by transiting the nodes with high probabilities in the variable space to improve the predicted response values of the ML model. The dataset was generated from three, 3D normal distributions to ensure that straight paths were not always actionable (Supplementary Fig. 2). Each distribution generated 200 data points that consisted of $x_1$, $x_2$, and $x_3$. The response variables were set to the sum of $x_1$, $x_2$, and $x_3$ with Gaussian noise ($\sigma = 2$). The dataset consisted of 600 data points and randomly split into training data (80%) and test data (20%).

### IHPP dataset

To evaluate our framework, we used the IHPP dataset. In this study, we considered the use case to plan actionable paths to improve SBP, which is a risk factor for cardiovascular disease[28].

The IHPP has annually acquired a wide range of health checkup data that comprise the molecular biology, physiology, biochemistry, lifestyle, and the socio-environmental aspects of residents of the Iwaki district, Hirosaki City, Aomori Prefecture, Japan. In this study, we targeted 12,803 health checkup instances with SBP values for 13 years (2005 to 2017). Given the existence of cases where the same person participated in multiple years, the number of unique participants was 3,132 (Table 1). The dataset was randomly split into training (80%) and test data (20%). This study was approved by the Ethics Committee of Hirosaki University School of Medicine (approval number: 2019–009) and was conducted according to the recommendations of the Declaration of Helsinki. All participants provided written informed consent.

### Construction of regression model

In our experiments, XGBoost[25], which is based on a gradient-boosting decision tree algorithm, was used to create the regression model. In general, XGBoost is a high-performance, nonlinear model. The hyperparameters of the model were determined by five-fold cross-validation of the training data. For preprocessing, continuous explanatory variables were standardized by the mean and standard deviation. In addition to this preprocessing, techniques such as other data-dependent preprocessing were applied. These data-dependent preprocessing is described in the experimental section.

## Surrogate model with hierarchical Bayesian modeling

The graphical model representation of the surrogate model constructed in this study is shown in Fig. 2. In this case, $\boldsymbol{x}_{cont}$ represents continuous explanatory variables, such as body composition and blood test data, $\boldsymbol{x}_{disc}$ represents discrete explanatory variables, such as sex, and $y$ represents a response variable, such as the blood pressure value. We denote the measured explanatory values as $\boldsymbol{x}$ and the predictions by the regression model as $y$. $\boldsymbol{z}$ is the parameter of the mixture components, and $k$ represents each mixture component. The data generative process is formulated as follows

$$\boldsymbol{x}_{cont} \sim N(\boldsymbol{m}_k, \boldsymbol{\Sigma}_k) \tag{1}$$

$$\boldsymbol{x}_{disc} \sim Categorical(\boldsymbol{\phi}_{disc,k}) \tag{2}$$

$$y \sim N(\mu_k, \sigma) \tag{3}$$

$$k \sim Categorical(\boldsymbol{\pi}) \tag{4}$$

$$\mu_k = \beta_{1,k} + \boldsymbol{\beta}_{2,k}^T \boldsymbol{x}_{cont} + \boldsymbol{\beta}_{3,k}^T \boldsymbol{x}_{disc} \tag{5}$$

$$\sigma = \frac{RMSE_{test}}{2} \tag{6}$$

where $RMSE_{test}$ is a root-mean-squared error of the regression model. The priors for other hyperparameters in equations (1)–(5) are defined as follows

$$\beta_{1,k} \sim N(y_{mean}, 5y_{std}) \tag{7}$$

$$\boldsymbol{\beta}_{2,k} \sim DoubleExponential(0, 1) \tag{8}$$

$$\boldsymbol{\beta}_{3,k} \sim DoubleExponential(0, 1) \tag{9}$$

$$\boldsymbol{\pi} \sim Dirichlet(\boldsymbol{1}) \tag{10}$$

$$\boldsymbol{m}_k \sim N(0, 5\boldsymbol{I}) \tag{11}$$

$$\boldsymbol{\Sigma}_k \sim diag(Cauchy(0, 2.5)) \tag{12}$$

$$\boldsymbol{\phi}_{disc,k} \sim Dirichlet(\boldsymbol{1}) \tag{13}$$

where $y_{mean}$ and $y_{std}$ represent the mean and standard deviation values of the predicted response variable, respectively. $\boldsymbol{\Sigma}_k$ is a diagonal matrix with elements according to the Cauchy distribution.

We used PyStan[46] to estimate the parameters using the Markov chain Monte Carlo algorithm (iteration = 1,500, warm-up = 500). We set 1–8 as the range of mixture components. For model selection, the WBIC was calculated for each model[24]. For a stable training, instances with values outside the 3σ range calculated using the training data in the continuous explanatory variables were excluded as outliers.

## Path planning using surrogate model

We calculated an optimal path for the treatment for each instance based on a breadth-first search algorithm. The intervention variable space was regarded as a grid graph, and the grid points (nodes) were connected to plan a path. The pseudocode of this algorithm is shown in Fig. 3. The purpose of this algorithm was used to obtain the most actionable path to the node that achieved the most improved predictive value within the search iteration count, *L*. From the computational perspective, we used the negative logarithm of actionability, defined as the product of node probabilities on a path, as a cost of the path. The goal of path planning is to discover the minimal cost path to each node. We obtained a list of nodes adjacent to the currently selected node in line 3 of the pseudocode. Subsequently, the costs for these nodes were updated in lines 5–7. The following node was selected in line 11. After reaching the predetermined count *L*, the path to the node with the best regression model prediction value was selected as the optimal path in line 13. If multiple nodes with the same predictions existed, the path with the minimum cost was selected.

**Actionability score**

To evaluate the actionability of the paths, we defined the actionability score expressed by the following equation: log(optimal path actionability) - log(baseline path actionability), where the optimal path actionability is the actionability of the path planned using our framework. The baseline path actionability is the geometric mean of 10 actionabilities of paths that connect both ends of the optimal path by the shortest procedure in a random manner. The actionability score indicated the actionability of the optimal path compared with that of baseline paths. When the actionability score is zero, the optimal path has the same actionability as the baseline path.

# Data availability

All data in this study are included in this article or are available from the corresponding author upon reasonable request.

# Code availability

All codes used in this study are available from the corresponding author upon a reasonable request.

## Acknowledgement

This work was supported by JST, Center of Innovation Program (JPMJCE1302), and Kyowa Hakko Bio Co., Ltd. The author, Kazuki Nakamura, thanks Kazushi Shoji, Miho Komatsu, Takashi Ishida, and Yuko Shimanami, employees of Kyowa Hakko Bio Co., Ltd., for their generous support.

## Authors information

**Affiliations**

**Research & Business Development Department, Kyowa Hakko Bio Co., Ltd., Tokyo, Japan**
Kazuki Nakamura

**Department of Biomedical Data Intelligence, Graduate School of Medicine, Kyoto University, Kyoto, Japan**

Kazuki Nakamura, Ryosuke Kojima, Eiichiro Uchino and Yasushi Okuno

**Center of Innovation Research Initiatives Organization, Hirosaki University, Zaifu-cho, Hirosaki, Aomori, Japan**
Koichi Murashita

**Department of Stress Response Science, Hirosaki University Graduate School of Medicine, Zaifu-cho, Hirosaki, Aomori, Japan**
Ken Itoh

**Department of Social Health, Hirosaki University Graduate School of Medicine, Zaifu-cho, Hirosaki, Aomori, Japan**
Shigeyuki Nakaji

## Contributions
K.N. and R.K. conceived of the presented idea. K.N. contributed to analyze data and draft the manuscript. R.K., E.U. and Y.O. provided substantial contributions to analyze data. K.M., K.I. and S.N. designed the study for acquisition of data. All authors critically interpret the results, reviewed and approved the manuscript.

## Corresponding authors
Correspondence to Yasushi Okuno.

# Ethics declarations

## Competing interests
Kazuki Nakamura is an employee of Kyowa Hakko Bio Co., Ltd. Other authors declare that they have no conflict of interest.

# Figures

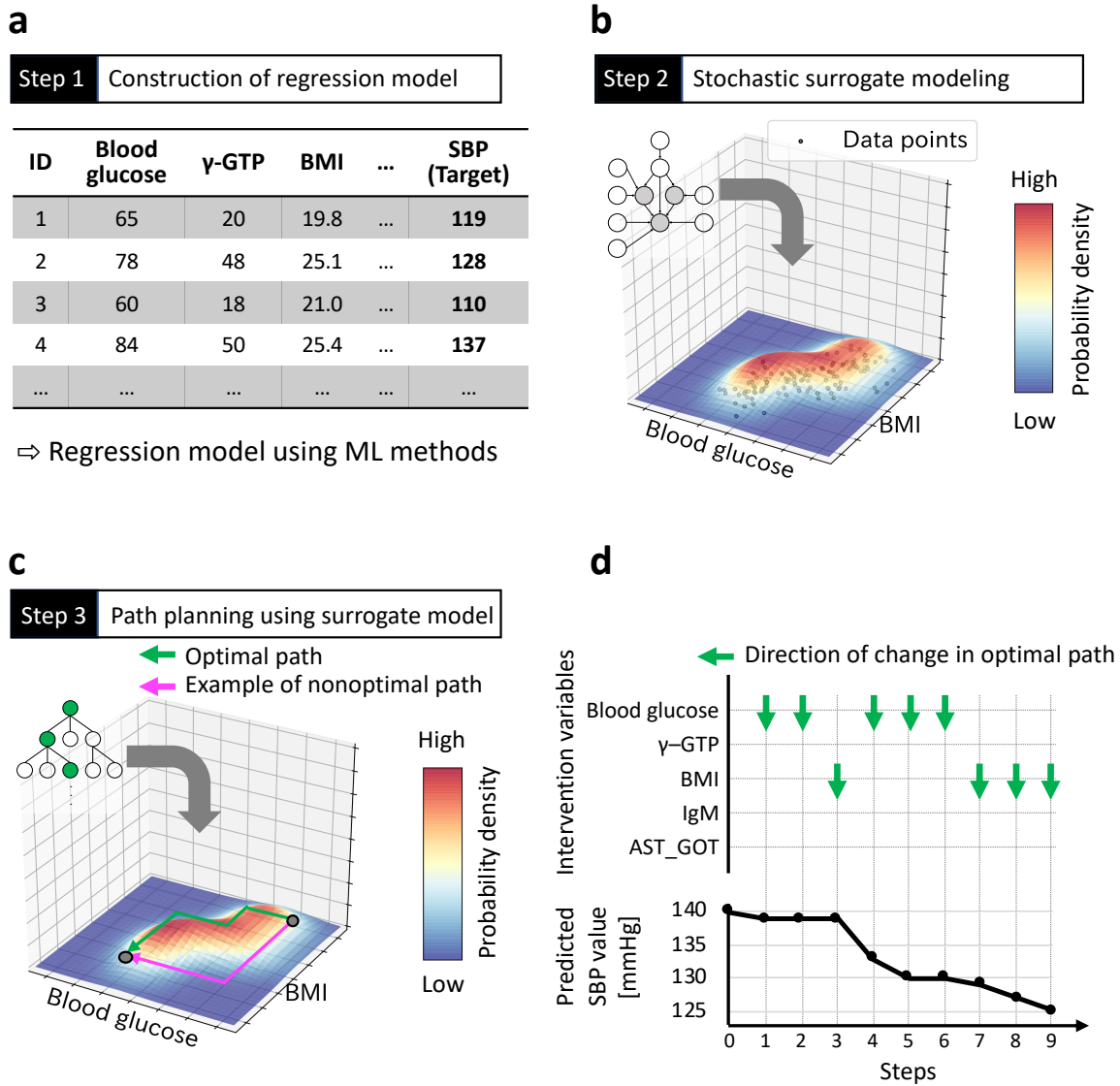

**Figure 1. Schematic representation of the framework for planning actionable paths for treatment using hierarchical Bayesian modeling.** The framework consists of three steps. A schematic is given as an example in which a path is planned to improve the systolic blood pressure (SBP) owing to changes in blood data and body composition data. **a** Construction of a regression model from the dataset. A variable SBP is set as the response variable in this case. **b** Construction of a stochastic surrogate model based on the original dataset and the predicted values of the regression model. This figure shows a schematic representation of a two-variable space regarding blood glucose and body mass index (BMI). The heatmap and vertical axis represent the existence probability of data in the variable space, which is expressed by the stochastic surrogate model. **c**, **d** Actionable path planning is applied to improve the response variable. The path is represented as a set of multistep transitions on explanatory variables. In our framework, the optimal path (green line in (**c**)) is planned on the grid graph with high probabilities in the variable space based on the stochastic surrogate model. Conversely, the nonoptimal path (red line in (**c**)) may pass through nodes with low or zero probability.

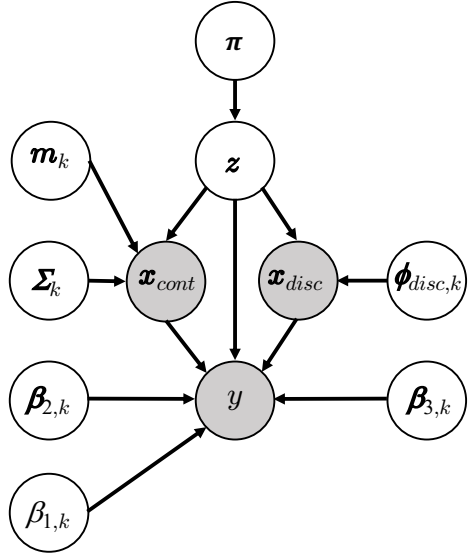

$$x_{cont} \sim N(m_k, \Sigma_k)$$
$$x_{disc} \sim Categorical(\phi_{disc,k})$$
$$y \sim N(\mu_k, \sigma)$$
$$k \sim Categorical(\pi)$$
$$\mu_k = \beta_{1,k} + \beta_{2,k}^T x_{cont} + \beta_{3,k}^T x_{disc}$$
$$\sigma = \frac{RMSE_{test}}{2}$$

$$\beta_{1,k} \sim N(y_{mean}, 5y_{std})$$
$$\beta_{2,k} \sim DoubleExponential(0,1)$$
$$\beta_{3,k} \sim DoubleExponential(0,1)$$
$$\pi \sim Dirichlet(\mathbf{1})$$
$$m_k \sim N(0, 5\mathbf{I})$$
$$\Sigma_k \sim diag(Cauchy(0, 2.5))$$
$$\phi_{disc,k} \sim Dirichlet(\mathbf{1})$$

**Figure 2. Graphical model representation of stochastic surrogate model.** Nodes in the graphical model are represented as follows: $x_{cont}$, continuous explanatory variables; $x_{disc}$, discrete explanatory variables; $y$, response variable predicted by the regression model; $z$, the parameter of the mixture components; and all the others, prior distributions. $k$ represents each mixture component, and $\Sigma_k$ is a diagonal matrix with elements according to the Cauchy distribution. The symbol $RMSE_{test}$ in the equation represents a root-mean-squared error of the regression model, and $y_{mean}$ and $y_{std}$ represent the mean and standard deviation values of the predicted response variable, respectively.

**Algorithm 1** Path search algorithm
**function** PATH_SEARCH
1: $current\_node \leftarrow$ initial node of instance
2: **for** $i = 1$ to $L$ **do**
3:     $neighbors \leftarrow$ list of node adjacent to $current\_node$
4:     **for** $neighbor\_node$ in $neighbors$ **do**
5:        $cost = current\_node.cost + neighbor\_node.neg\_log\_prob$
6:        **if** $neighbor\_node.cost > cost$ **then**
7:           $neighbor\_node.cost = cost$
8:        **end if**
9:     **end for**
10:    $current\_node.visited =$ True
11:    $current\_node \leftarrow$ node which is not $visited$ and has smallest $cost$
12: **end for**
13: $destination\_node \leftarrow$ node with the best response variable
14: **return** $destination\_node$

**Figure 3. Pseudocode of path search algorithm.**

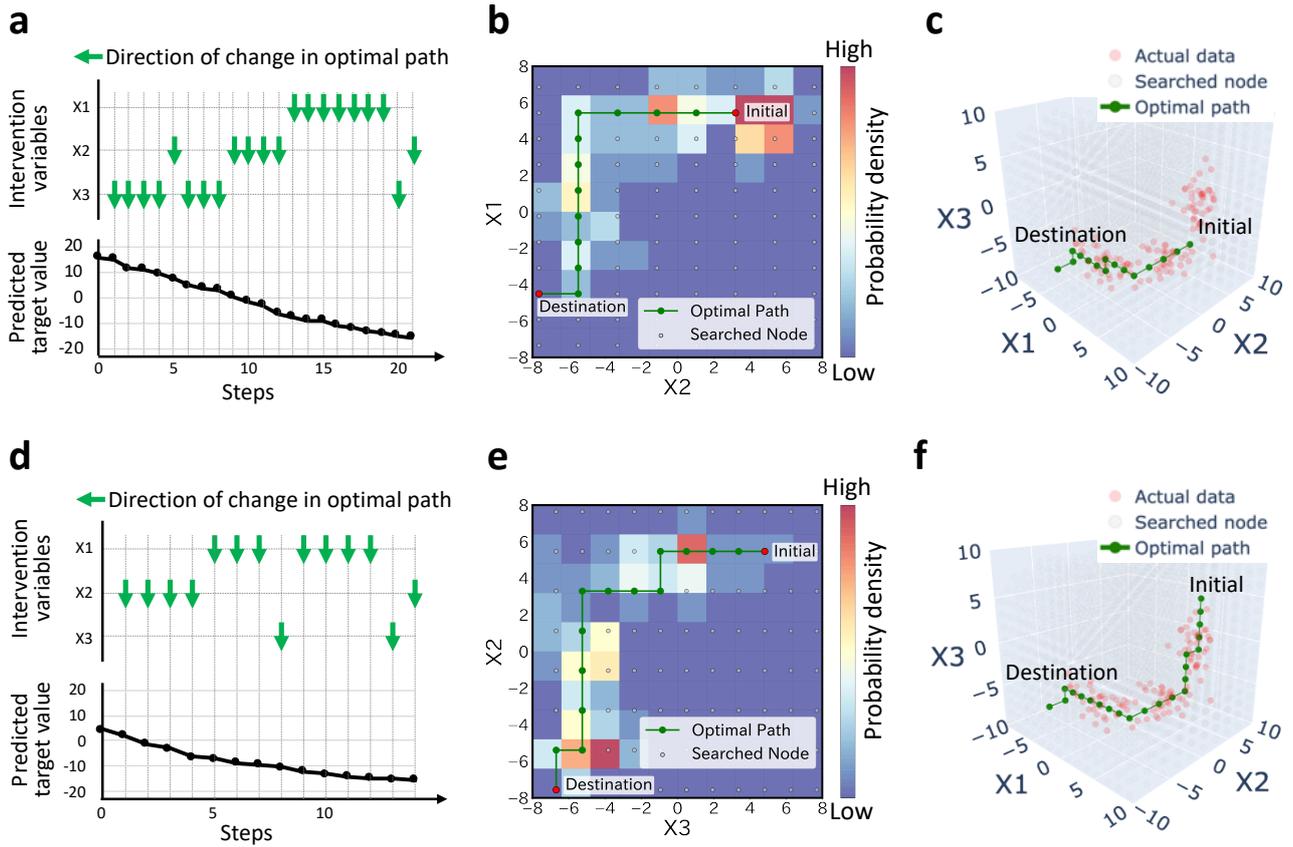

**Figure 4. Examples of actionable paths planned on synthetic dataset.** The optimal paths for improving the response variable predicted by the ML model are represented for randomly selected two examples: instance 1 (**a**–**c**) and instance 2 (**d-f**). **a**, **d** The orders of changes in the explanatory variables in the optimal path and the accompanying changes in the predicted values. In the transition steps, the upward or downward arrow represents a unit increase or decrease in the explanatory variable, respectively. **b**, **e** Two-dimensional (2D) plots of the path. The 2D plots are shown regarding the selected two variables: X1 and X2 (**b**), and X2 and X3 (**e**). **c**, **f** Three-dimensional (3D) plots of the path.

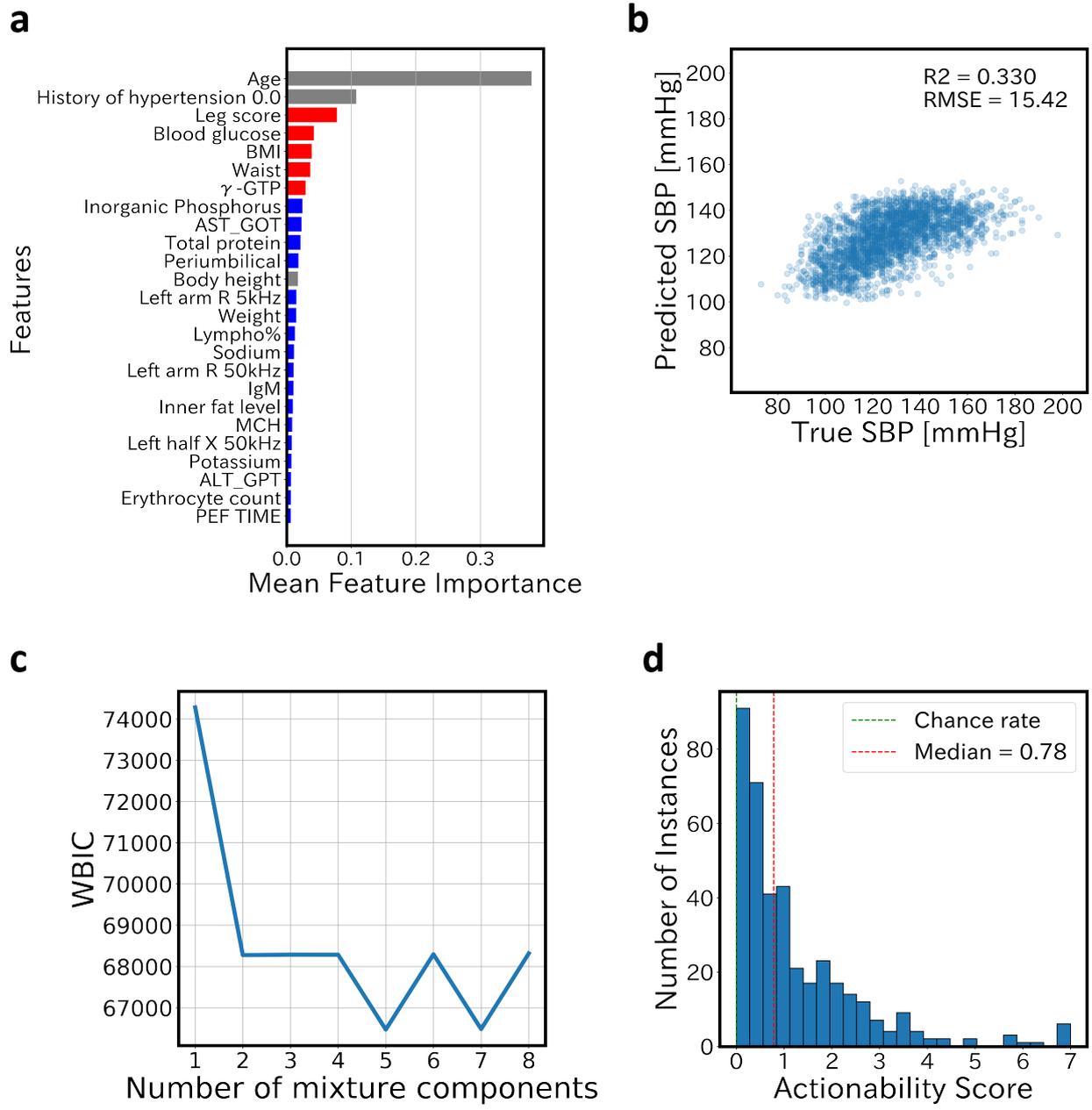

**Figure 5. Results of proposed framework on the Iwaki Health Promotion Project (IHPP) dataset. a** Mean feature importance: these 25 features were selected by recursive feature elimination (RFE) to predict the systolic blood pressure (SBP). RFE was performed with five-fold cross-validation, and the importance of all the mean features was calculated when 25 variables remained. The color of each bar represents the following: red: intervention variables in path planning, gray: variables which cannot be easily intervened, and blue: other variables. Details of features are described in Supplementary Table 1. **b** Plot for prediction vs. true response variable. **c** Widely applicable Bayesian information criterion (WBIC) values of stochastic surrogate models with 1–8 mixture components. **d** Histogram of actionability scores at different instances. An actionability score of zero indicates that the actionability of the optimal path is equivalent to that of the baseline path. Only five instances scored over seven, with a maximum score >14. Scores of these instances were summarized in a score of seven to adjust the plot appearance.

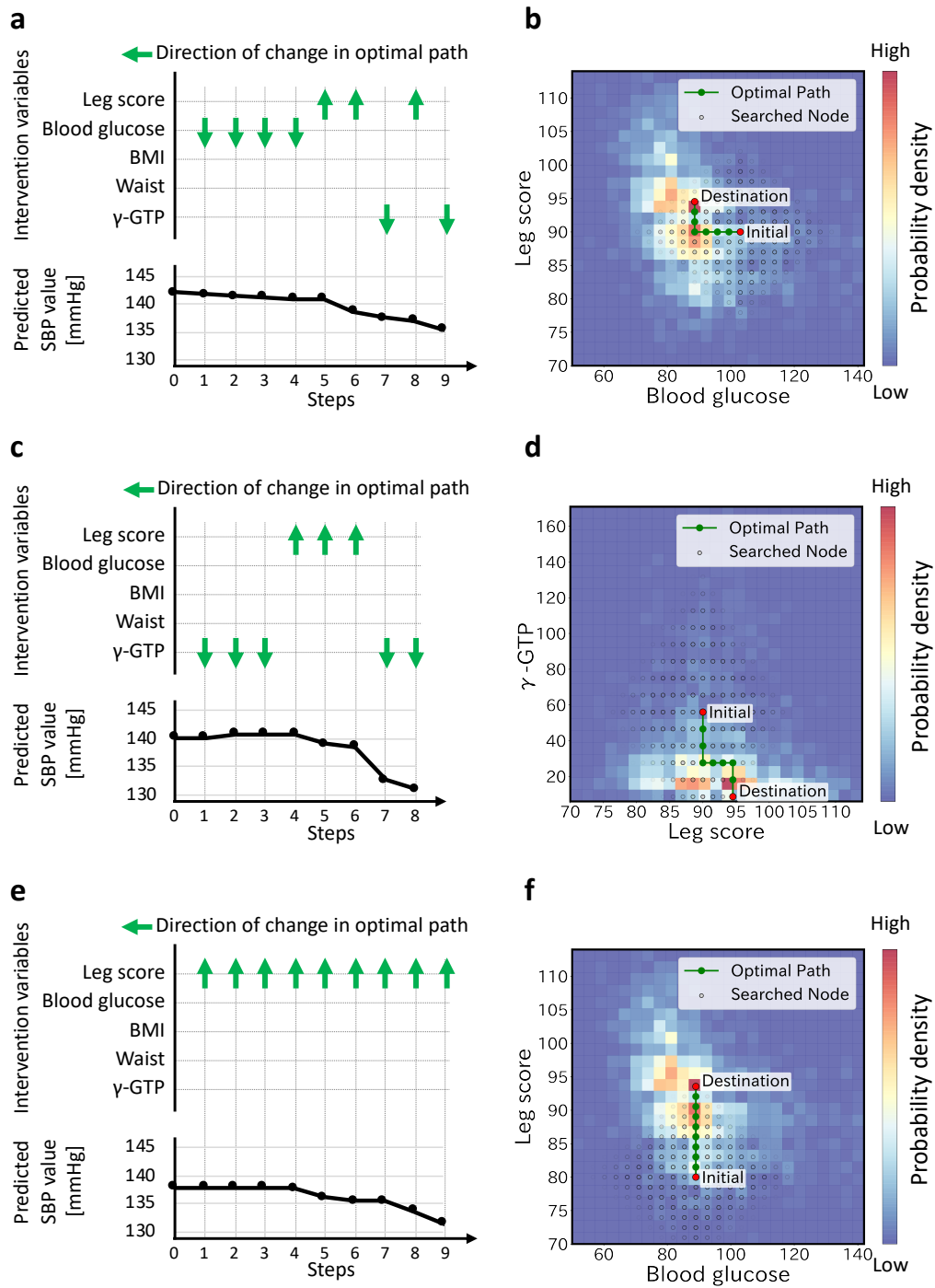

**Figure 6. Examples of personal actionable paths for treatment using the IHPP dataset.** The optimal paths for improving the response variable predicted by the ML model are represented for randomly selected three examples: instance 1 (**a**, **b**), instance 2 (**c**, **d**), and instance 3 (**e**, **f**). **a**, **c**, **e** The orders of changes in the explanatory variables in the optimal path and the accompanying changes in the predicted values. In the transition steps, the upward or downward arrow represents a unit increase or decrease in the explanatory variable, respectively. **b**, **d**, **f** 2D plots of the path. The 2D plots are shown regarding the two influential variables in the optimal path: blood glucose and leg score (**b**), leg score and γ-GTP (**d**), and blood glucose and leg score (**f**). 3D plots of the path are shown in Supplementary Fig. 4.

# Tables

Table 1. Subject characteristics during first-time participation.

| Participant characteristics | (N = 3132) |
|---|---|
| Age (years) | 51.3 ± 16.0 |
| BMI (kg/m$^2$) | 23.0 ± 3.5 |
| SBP (mmHg) | 125.8 ± 19.1 |
| DBP (mmHg) | 74.8 ± 11.9 |
| Sex | |
|     Male | 1234 (39.4%) |
|     Female | 1898 (60.6%) |
| History of hypertension | |
|     No history | 2445 (78.1%) |
|     Under treatment | 651 (20.8%) |
|     Cured | 36 (1.1%) |

BMI: Body Mass Index, SBP: Systolic Blood Pressure, DBP: Diastolic Blood Pressure.

# Supplementary Information

# Health improvement framework for planning actionable treatment process using surrogate Bayesian model

Kazuki Nakamura[1,2], Ryosuke Kojima[2], Eiichiro Uchino[2], Koichi Murashita[3], Ken Itoh[4], Shigeyuki Nakaji[5] and Yasushi Okuno[2*]

[1] Research & Business Development Department, Kyowa Hakko Bio Co., Ltd., Tokyo, Japan.

[2] Department of Biomedical Data Intelligence, Graduate School of Medicine, Kyoto University, Kyoto, Japan.

[3] Center of Innovation Research Initiatives Organization, Hirosaki University, Zaifu-cho, Hirosaki, Aomori, Japan.

[4] Department of Stress Response Science, Hirosaki University Graduate School of Medicine, Zaifu-cho, Hirosaki, Aomori, Japan.

[5] Department of Social Health, Hirosaki University Graduate School of Medicine, Zaifu-cho, Hirosaki, Aomori, Japan.

## Supplementary experiments

Because the regression model based on the health dataset used in the main text had a low-prediction score, we also applied our framework on a non-health dataset and another simple health dataset.

### Evaluation on real estate dataset

We obtained a dataset from the UCI Machine Learning Repository[1,2]. This dataset was used to conduct a regression analysis of the house price of a unit area based on six continuous explanatory variables of real estate data. There were no missing values in the dataset. The dataset was randomly split into training data (80%) and test data (20%).

The trained regression model yielded a root-mean-square error (RMSE) of 6.78 and an R-squared value of 0.735 in the test data (Supplementary Fig. 5a). A surrogate model was constructed by hierarchical Bayesian modeling using the original data and the predicted values of the regression model. The lowest WBIC value was obtained when the number of mixture components was two (Supplementary Fig. 5b).

Subsequently, path planning was performed using the surrogate model. Among the six explanatory variables, X1 (the transaction date) was a variable that was difficult to intervene. Therefore, X1 was fixed, and the remaining five variables were selected as intervention variables. The unit cell size of the grid was set to 0.2 σ in the training data for each explanatory variable. We executed the path search algorithm with $L = 20,000$ for each instance, and the path with the highest predicted value was acquired. The histogram of the actionability score for each instance is shown in Supplementary Fig. 5c. The actionability scores were greater than zero in 75/82 instances, and the median was 5.25.

From these results, we have demonstrated that our framework can be applicable to datasets with higher regression model scores.

**Evaluation on public dataset for disease progression**

To evaluate the feasibility of our framework on another small health dataset, we used a public dataset on diabetes progression[3,4]. This dataset was used to conduct a regression of the quantitative measure of diabetes progression one year after the baseline from nine continuous and one discrete explanatory variable (Supplementary Table 2). The dataset is openly available on Trevor Hastie's Software page at https://web.stanford.edu/~hastie/Papers/LARS/. This dataset contains no missing values. The dataset was randomly split into training data (80%) and test data (20%).

The feature importance of the trained model is shown in Supplementary Fig. 6a. The regression model yielded an RMSE value of 62.19 and an R-squared value of 0.246 in the test data (Supplementary Fig. 6b). A surrogate model was constructed by hierarchical Bayesian modeling using the original data and the predicted values of the regression model. The lowest WBIC value was obtained when the number of mixture components was two (Supplementary Fig. 6c).

Subsequently, path planning was performed using the surrogate model. Regarding the intervention variables, five variables were selected from the top of the feature importance of the regression model: body mass index (BMI), blood pressure, T-cells, high-density lipoproteins, and lamotrigine (Supplementary Fig. 6a). The remaining variables were fixed. The unit cell size of the grid was set to 0.2 σ in the training data for each explanatory variable. We executed the path search algorithm with $L = 20,000$ in each instance, and the path with the lowest predicted value was acquired. The histogram of the actionability score for each instance is shown in Supplementary Fig. 6d. The actionability scores were greater than zero in 83/87 instances, and the median was 2.06. Examples of the paths planned by using the proposed framework are shown in Supplementary Fig. 7. This experiment indicates the feasibility of using the proposed framework in planning actionable paths to improve the predictions of the regression model.

407–499 (2004).

4. Hastie, T. & Efron, B. lars: Least Angle Regression, Lasso and Forward Stagewise. *R package version 1.2* (2013).

# Supplementary Figures

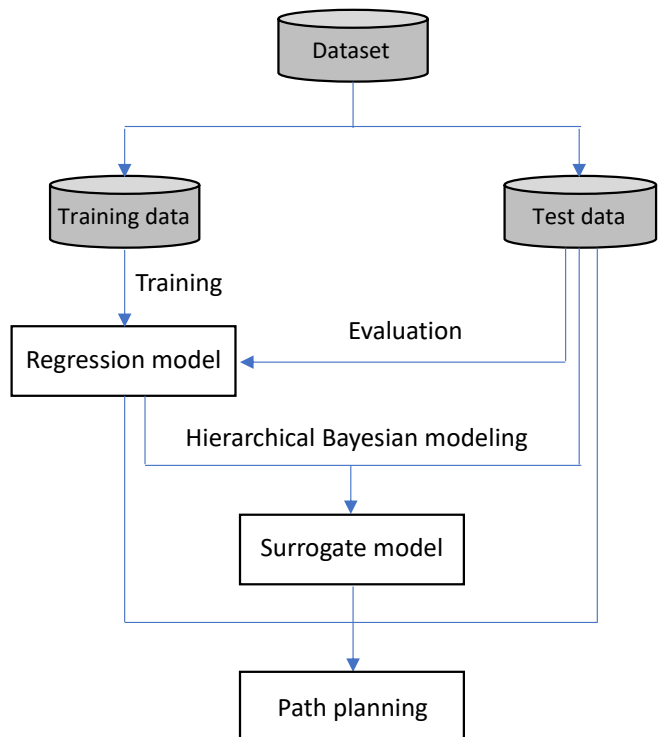

**Supplementary Figure 1. Detailed workflow of proposed framework.**

$$\begin{bmatrix} x_1 \\ x_2 \\ x_3 \end{bmatrix} \sim \mathcal{N}(\boldsymbol{\mu}, \boldsymbol{\Sigma})$$

$$\boldsymbol{\mu}_1 = \begin{bmatrix} 0 \\ -5 \\ -5 \end{bmatrix} \quad \boldsymbol{\Sigma}_1 = \begin{bmatrix} 5 & 0 & 0 \\ 0 & 1 & 0 \\ 0 & 0 & 1 \end{bmatrix}$$

$$\boldsymbol{\mu}_2 = \begin{bmatrix} 5 \\ 0 \\ -5 \end{bmatrix} \quad \boldsymbol{\Sigma}_2 = \begin{bmatrix} 1 & 0 & 0 \\ 0 & 5 & 0 \\ 0 & 0 & 1 \end{bmatrix}$$

$$\boldsymbol{\mu}_3 = \begin{bmatrix} 5 \\ 5 \\ 0 \end{bmatrix} \quad \boldsymbol{\Sigma}_3 = \begin{bmatrix} 1 & 0 & 0 \\ 0 & 1 & 0 \\ 0 & 0 & 5 \end{bmatrix}$$

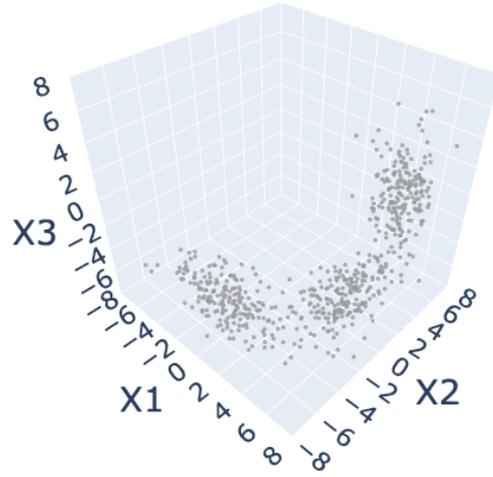

**Supplementary Figure 2. Generation of synthetic dataset.** The three, three-dimensional (3D) normal distributions generated 200 data points that consisted of $x_1$, $x_2$, and $x_3$. Subsequently, a response variable was set to the sum of $x_1$, $x_2$, and $x_3$ with Gaussian noise ($\sigma = 2$). The synthetic dataset consisted of a total of 600 data points with explanatory variables (X1, X2, and X3) and a response variable.

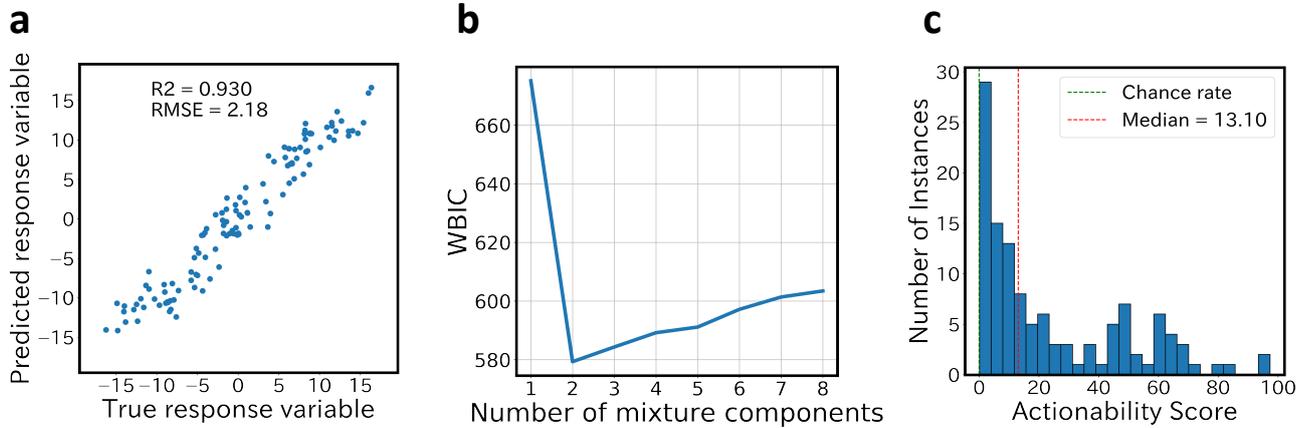

**Supplementary Figure 3. Results of the proposed framework on synthetic dataset. a** Plot for prediction vs. true response variable. **b** Widely applicable Bayesian information criterion (WBIC) values of the stochastic surrogate models with 1–8 mixture components. **c** Histogram of actionability scores at different instances. The unit cell size of the grid was set to 0.5 $\sigma$ in the training data for each explanatory variable. The path search algorithm was executed with $L = 20{,}000$ for each instance and acquired a path with the lowest predictive value. An actionability score of zero indicates that the actionability of the optimal path is equivalent to that of the baseline path.

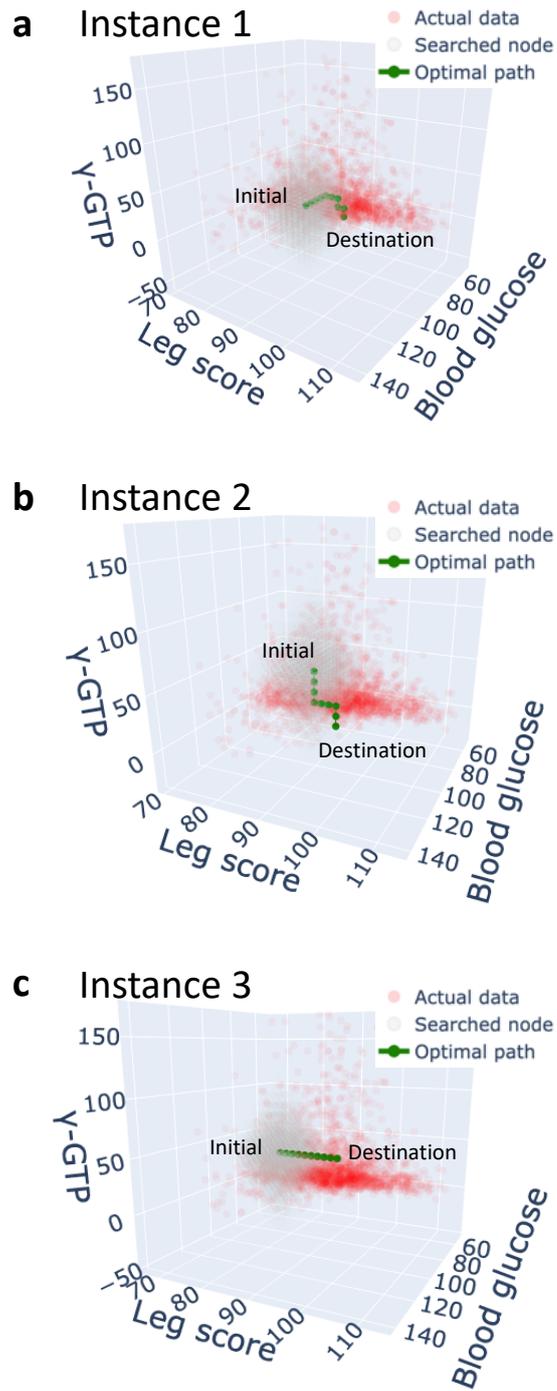

**Supplementary Figure 4. Plots of optimal paths in 3D for the Iwaki Health Promotion Project (IHPP) dataset.**
The instances respectively correspond to the instances in Fig. 6 of the main text.

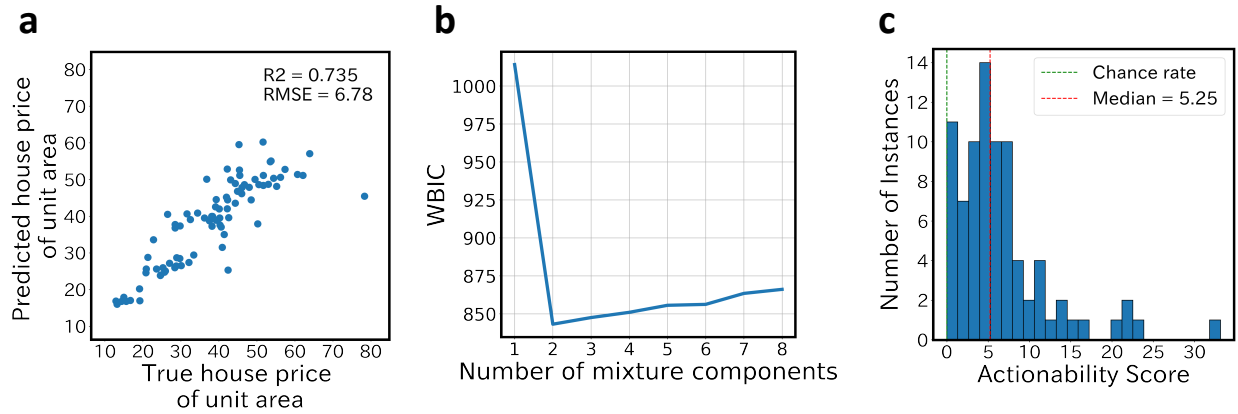

**Supplementary Figure 5. Results of proposed framework on real estate dataset. a** Plot for prediction vs. true response variable. **b** WBIC values of the stochastic surrogate models with 1–8 mixture components. **c** Histogram of actionability scores at different instances. An actionability score of zero indicates that the actionability of the optimal path is equivalent to that of the baseline path.

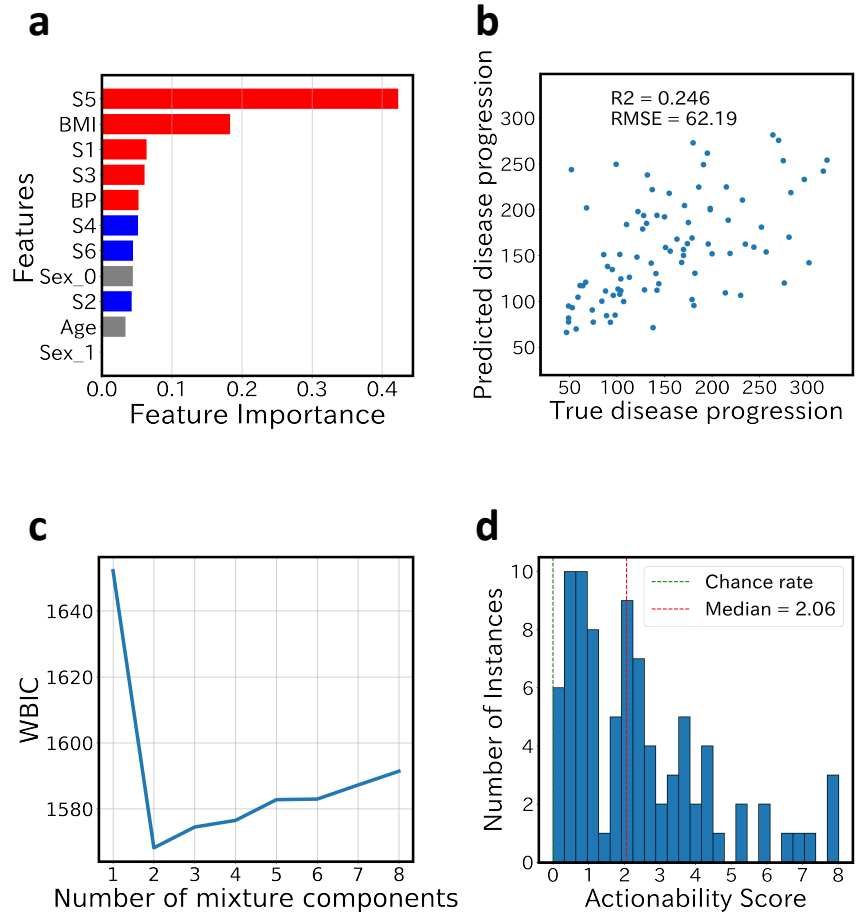

**Supplementary Figure 6. Results of proposed framework on diabetes progression dataset. a** Feature importance of the regression model. The color of each bar represents the following: red: intervention variables in path planning, gray: variables which are difficult to be intervened, and blue: other variables. **b** Plot for prediction vs. true response variable. **c** WBIC values of stochastic surrogate models with 1–8 mixture components. **d** Histogram of actionability scores for each instance. An actionability score of zero indicates that the actionability of the optimal path is equivalent to that of the baseline path.

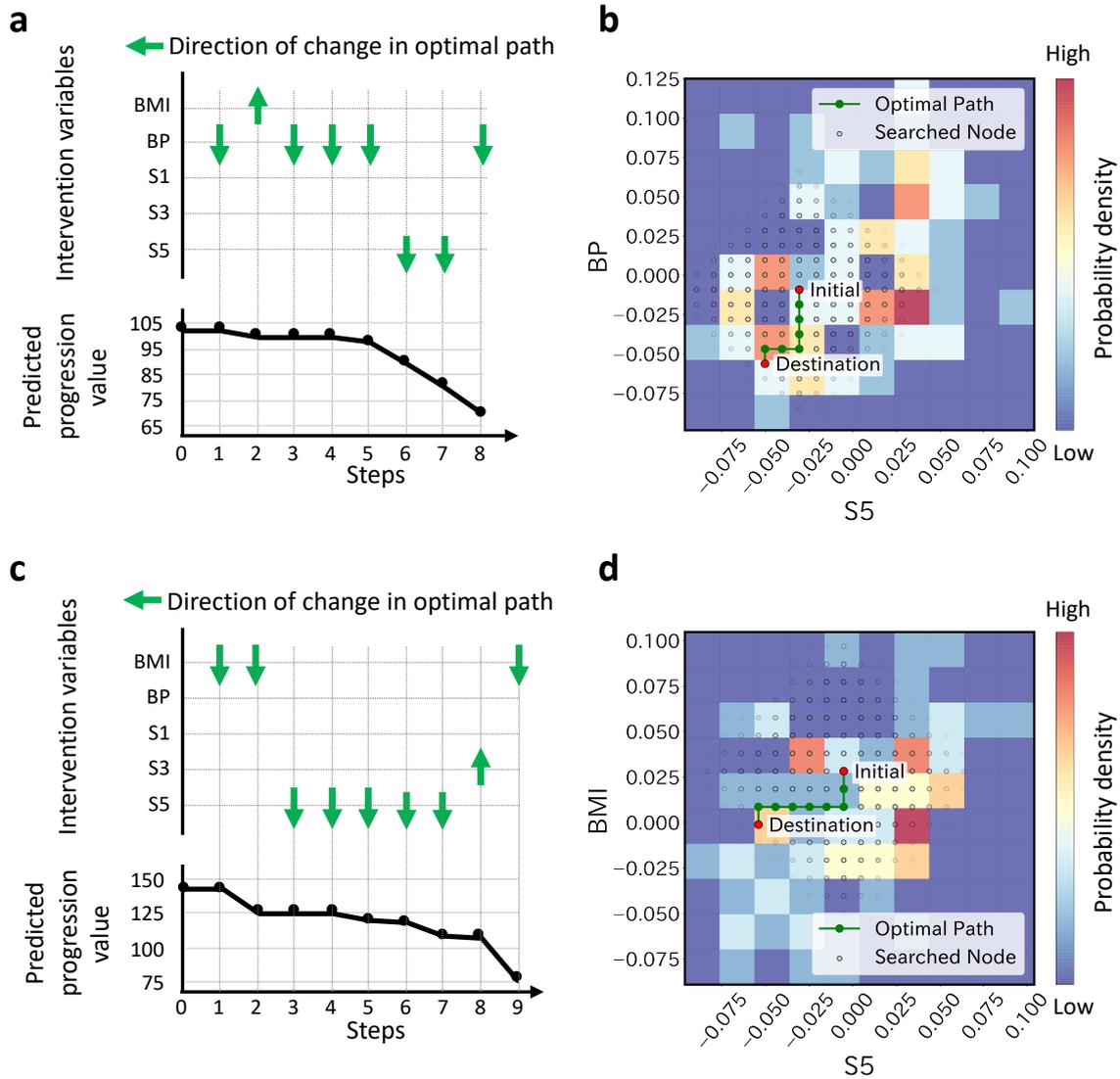

**Supplementary Figure 7. Examples of personal actionable paths for treatment on diabetes progression dataset.** The optimal paths needed for the improvement of the response variables predicted by the machine learning model are represented for randomly selected two examples: instance 1 (**a**, **b**) and instance 2 (**c**, **d**). **a**, **c** The orders of changes in the explanatory variables in the optimal path and the accompanying changes in the predicted values. In the transition steps, the upward or downward arrow represents a unit increase or decrease in the explanatory variable, respectively. **b**, **d** Two-dimensional (2D) plots of the path. The 2D plots are shown regarding the two influential variables: S5 and average blood pressure (BP) (**b**), and S5 and body mass index (BMI) (**d**).

# Supplementary Tables

**Supplementary Table 1. Description of the recursive feature elimination (RFE)-selected features on the Iwaki Health Promotion Project (IHPP) dataset.**

| Features | Description |
| --- | --- |
| Age (years) | |
| History of hypertension 0.0 | One-hot vector which represents no history of hypertension |
| Leg score | Leg skeletal muscle level score calculated by a body composition meter |
| Blood glucose (mg/dL) | |
| BMI (kg/m$^2$) | Body mass index |
| Waist (cm) | Waist circumference |
| γ-GTP (U/L) | |
| Inorganic phosphorus (mg/dL) | Serum inorganic phosphorus |
| AST_GOT (U/L) | Aspartate transaminase |
| Total protein (g/dL) | |
| Periumbilical (cm) | Circumference of navel |
| Body height (cm) | |
| Left arm R 5 kHz | Bioelectrical resistance parameter of left arm measured by a body composition meter |
| Weight (kg) | |
| Lymphocytes (%) | Percentage of lymphocytes in white blood cells |
| Sodium (mEq/L) | Serum sodium |
| Left arm R 50 kHz | Bioelectrical resistance parameter of left arm measured by body composition meter |
| IgM (mg/dL) | Immunoglobulin M |
| Inner fat level | Inner fat level calculated by body composition meter |
| MCH (pg) | Mean corpuscular hemoglobin |
| Left half X 50 kHz | Bioelectrical reactance parameter of left half of the body measured by body composition meter |
| Potassium (mEq/L) | Serum Potassium |
| ALT_GPT (U/L) | |
| Erythrocyte count (×10$^4$/μL) | Erythrocyte count in blood |
| PEF TIME | Peak expiratory flow–time from spirometry |

**Supplementary Table 2. Description of the diabetes progression dataset features.**

| Feature | Description |
| --- | --- |
| Response variable | |
|     Disease progression | A quantitative measure of disease progression one year after baseline |
| Explanatory variables | |
|     Age | Age in years |
|     Sex | |
|     BMI | Body mass index |
|     BP | Average blood pressure |
|     S1 | T-cells (a type of white blood cells) |
|     S2 | Low-density lipoproteins |
|     S3 | High-density lipoproteins |
|     S4 | Thyroid stimulating hormone |
|     S5 | Lamotrigine |
|     S6 | Blood sugar level |